\documentclass[10pt,twocolumn,letterpaper]{article}
\pdfoutput=1
\usepackage{times}
\usepackage{epsfig}
\usepackage{a4wide}
\usepackage{times}
\usepackage{url}
\usepackage{amsthm}
\usepackage{amsmath}
\usepackage{amssymb}
\usepackage{dsfont}
\usepackage{graphicx}
\usepackage{algorithm}
\usepackage{algpseudocode}
\usepackage[normalem]{ulem}
\usepackage{xcolor}

\usepackage[breaklinks=true,bookmarks=false]{hyperref}

\newcommand{\x}{x}
\newcommand{\y}{\boldsymbol{y}}
\renewcommand{\l}{l}

\newcommand{\MoRF}{\textrm{MoRF}}
\DeclareMathOperator{\area}{area}
\newcommand{\eps}{\epsilon}

\begin{document}
\title{Analyzing Classifiers: Fisher Vectors and Deep Neural Networks}
\author{Sebastian Bach\\
Fraunhofer HHI\\
Einsteinufer 37, 10587 Berlin\\
{\tt\small sebastian.bach@hhi.fraunhofer.de}
\and
Alexander Binder\\
SUTD\\
8 Somapah Rd, 487372 Singapore\\
{\tt\small alexander\_binder@sutd.edu.sg}
\and 
Gr\'egoire Montavon\\
TU Berlin\\
Marchstr. 23, 10587 Berlin. Germany\\
{\tt\small gregoire.montavon@tu-berlin.de}
\and 
Klaus-Robert M\"uller\\
TU Berlin\\
Marchstr. 23, 10587 Berlin. Germany\\
{\tt\small klaus-robert.mueller@tu-berlin.de}
\and 
Wojciech Samek\\
Fraunhofer HHI\\
Einsteinufer 37, 10587 Berlin\\
{\tt\small wojciech.samek@hhi.fraunhofer.de}
}
\date{}
\maketitle

\section{Abstract}
Fisher Vector classifiers and Deep Neural Networks (DNNs) are popular and successful algorithms for solving image classification problems. However, both are generally considered `black box' predictors as the non-linear transformations involved have so far prevented transparent and interpretable reasoning. Recently, a principled technique, Layer-wise Relevance Propagation (LRP), has been developed in order to better comprehend the inherent structured reasoning of complex nonlinear classification models such as Bag of Feature models or DNNs. In this paper we (1) extend the LRP framework also for Fisher Vector classifiers and then use it as
analysis tool to (2) quantify the importance of context for classification, (3) qualitatively compare DNNs against FV classifiers in terms of important image regions and (4) detect potential flaws and biases in data.  All experiments are performed on the PASCAL VOC 2007 data set.

\section{Introduction}
Deep neural networks have defined state of the art in many fields, such as image classification \cite{DBLP:conf/nips/KrizhevskySH12}, image detection \cite{DBLP:journals/corr/GirshickDDM13} and machine translation \cite{DBLP:conf/nips/SutskeverVL14}. While much of research is devoted to extending the applicability of deep neural nets to more domains \cite{Hochreiter:1997:LSM:1246443.1246450, DBLP:conf/gecco/KoutnikCSG13,DBLP:journals/corr/ZhangDGD14,DBLP:conf/cvpr/KarpathyTSLSF14,DBLP:conf/nips/KarpathyJL14}, we focus here on a different question, namely the impact of context, and the ability to use context. This question was raised already during times of the Pascal VOC challenge, where the amount of context was a matter of speculation, c.f.~PASCAL VOC workshop presentation slides in \cite{pascalvocworkshop2010}. 

The question of context is considered for two prominent types of classifiers. The first type, Fisher Vectors (FV) \cite{sanchez2013image} are based on computing a single feature map on an image as a whole and subsequently computing one score. In such a setup one can expect that context plays naturally a role for the prediction as the image is processed as a whole during training and test time.
In case of small training sample sizes and the absence of opportunities for fine-tuning, Fisher vectors still might be a viable alternative to deep neural nets due to their reduced parameter space. Examples for performance issues of deep neural networks on small sample sizes without finetuning can be seen in \cite{DBLP:conf/eccv/ZeilerF14}. 
The question of context is also open for the second type, Deep Neural Networks (DNN). One might assume that context plays no role for neural networks when they are used in classification by detection setups. For example, a recent Imagenet challenge winner relied on $144$ crops per test image and classifier \cite{DBLP:journals/corr/SzegedyLJSRAEVR14}. Another work using Pascal VOC data \cite{DBLP:conf/cvpr/OquabBLS14} used at test time $500$ multi-scale patches per test image. However in certain setups computing several hundred windows as required for classification by detection setups may not be possible, e.g. when using hardware without GPUs and much main memory, such as used consumer laptops or smartphones, and when having time constraints for computation of the test prediction on an image. One can expect to see a larger impact of context when resorting to a few regions of an image at test time only, and thus training and testing with larger image patches.

Our contribution here is as follows. (1) We extend the method of \cite{BacBinMonKlaMueSam15} to Fisher vectors, and apply relevance propagation for the first time to Fisher vectors. (2) We define measures for the amount of context used for prediction in a single test image. (3) We apply the measures of context for neural networks and Fisher vector based classifiers on the Pascal VOC dataset, as it offers a way to approximately validate context by its bounding box annotation. We compare the context dependence of Fisher vectors against neural nets which were trained on larger patches of input images. (4) We show that this methodology is able to identify strong cases of context and biases in the training data even without using bounding box information.

The next section reviews related work. Section \ref{sec:fisher} briefly describes the Fisher Vector classifier. Section \ref{sec:decomposing} introduces the extended LRP method to decompose a Fisher Vector prediction into scores for small regions of the order of a local feature. The same section also proposes a novel LRP-based measure of the importance of context. Section \ref{sec:experiments} introduces the experimental setup and presents results. The paper concludes in Section \ref{sec:discussion} with a summary and an outlook.

\section{Related Work}
In recent years, interest in understanding image representations \cite{DBLP:journals/corr/MahendranV14, DBLP:journals/corr/LencV14, DBLP:journals/corr/NovotnyLPV15} and being able to explain the decision process of a classification system has increased, with e.g., gradient-based sensitivity analysis \cite{DBLP:journals/jmlr/BaehrensSHKHM10, DBLP:journals/corr/SimonyanVZ13}. However, many approaches have been conceived with a specific pipeline architecture in mind. So do \cite{Uijlings12} explain predictions for bag of word features with hard mapping (Vector Quantization) and Histogram Intersection kernels, and \cite{liu2012has} identifies image regions critical for the prediction of a linear SVM classifier with max-pooling feature aggregation algorithm. A solution especially dedicated to visualize image regions triggering the prediction of deep convolutional neural networks with max-pooling layers and has been proposed in \cite{DBLP:conf/eccv/ZeilerF14} with \emph{deconvolution nets}. 

Recently, a paradigm called Layer-wise Relevance Propagation (LRP) has been introduced in \cite{BacBinMonKlaMueSam15} as a way to compute partial prediction contributions -- or \emph{relevance values} $R$ -- for intermediate and input representations based on the final classifier output. It computes scores for regions or pixels of an image explaining the prediction itself rather than the effect of single neurons or particular layers. It is applied in \cite{BacBinMonKlaMueSam15} to Bag of visual words classifiers and deep neural networks; in this paper we extend this method to make it applicable to Fisher Vector classifiers.

\section{Fisher Vectors in a Nutshell}
\label{sec:fisher}
\begin{figure*}[t]
\centering
\includegraphics[width=1\textwidth]{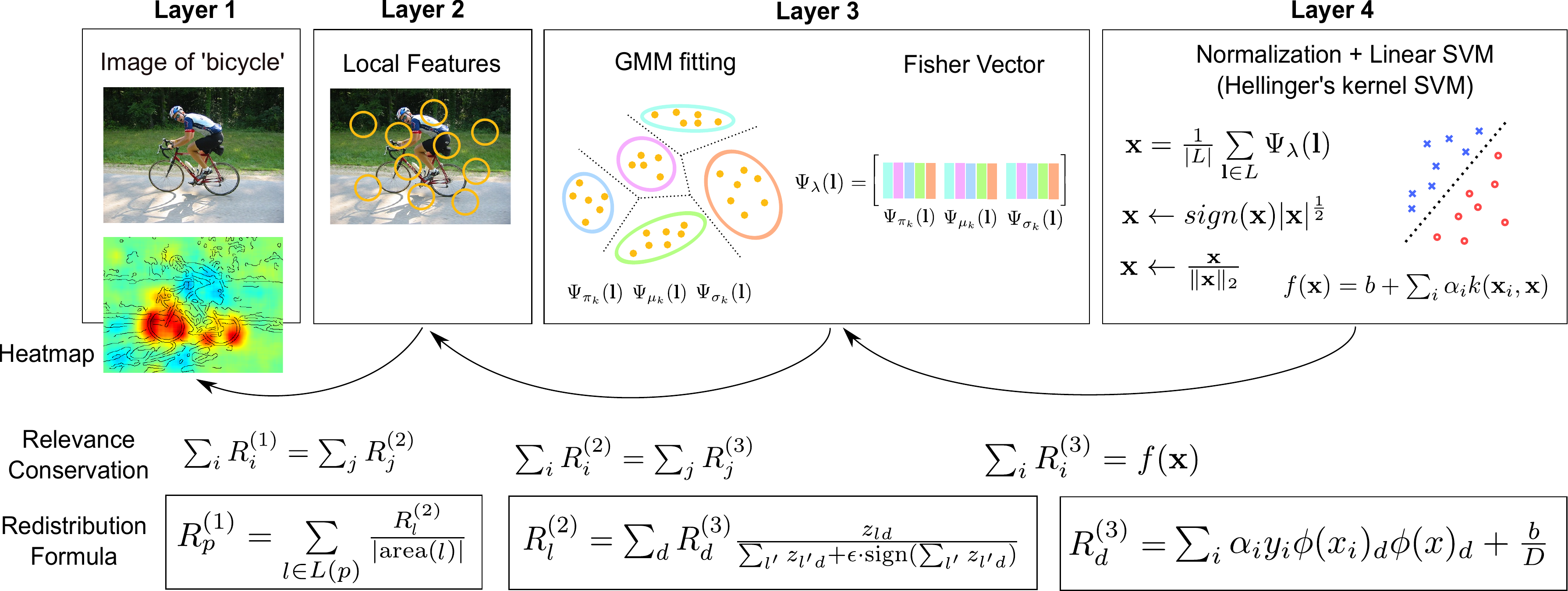}
\caption{Computing Fisher Vector representation of an image and explaining the classification decision.}
\label{fig:fv}
\end{figure*}

Fisher Vectors \cite{perronnin2010improving, sanchez2013image} are a powerful tool to compute rich image or video representations and provide state-of-the-art performance amongst feature extraction algorithms.
Figure \ref{fig:fv} summarizes the steps involved in computing FV representation of an image. We introduce here  a notation which later will be used in the Section \ref{sec:decomposing}.

An integral part for computing FVs is to fit a Gaussian Mixture Model (GMM) on top of the local descriptors $L = \{\l\}$ extracted from the training data to serve as a soft vocabulary of visual prototypes. Assuming a $K$-component GMM $\lambda = \lbrace (\pi_k, \mu_k, \Sigma_k) \rbrace_{k=1..K}$, then $\pi_k$ is the mixture weight of component $k$, with $\sum_k \pi_k = 1$ and $\forall k : \pi_k \geq 0$, $\mu_k$ is the mean vector of the $k$th mixture component and $\Sigma_k$ its (diagonal) covariance matrix. For the computation of a \emph{full} FV representation of an image, each local descriptor $l$ is related to all $K$ components of the trained GMM in its $0$th (soft mapping weight), $1$st (deviation from mean) and $2$nd moment (variance) \cite{sanchez2013image}:
\begin{align}
\Psi_{\pi_k}(\l) & = \frac{1}{\sqrt{\pi_k}}\left( \gamma_k(\l) - \pi_k \right) \\
\Psi_{\mu_k}(\l) & = \frac{1}{\sqrt{\pi_k}}\gamma_k(\l)\left( \frac{\l - \mu_k}{\sigma_k} \right)\\
\Psi_{\sigma_k}(\l)& = \frac{1}{\sqrt{\pi_k}}\gamma_k(\l)\frac{1}{\sqrt{2}}\left( \frac{\left(\l - \mu_k\right)^2}{\sigma_k^2} - 1 \right)
\end{align}
with $\Psi_{\pi_k}(\l) \in \mathbb{R}$ , both $\Psi_{\mu_k}(\l)$ and $\Psi_{\sigma_k}(\l) \in \mathbb{R}^D$ and $\gamma_k(\l)$ returning the soft assignment of $\l$ to the $k$th mixture component.
The FV embedding $\Psi_\lambda(\l)$ for a single descriptor $\l$ is then achieved by concatenating the mapping outputs relative to all $K$ components into a $(1+2D)K$ dimensional vector
\begin{align}
\Psi_\lambda(\l) = \left[\Psi_{\pi_1}(\l)\ldots \Psi_{\mu_1}(\l)\ldots \Psi_{\sigma_1}(\l)\ldots\right]
\label{eq:fvassemble}
\end{align}
Having computed all those (as we will refer to now as) \emph{raw} Fisher embeddings for all individual local descriptors, a single image-wise descriptor is achieved by averaging over the complete set of $\Psi_\lambda(\l)$, followed by power normalization to reduce the sparsity of the descriptor and $\ell_2$-normalization to improve prediction performance \cite{perronnin2010improving}. The application of both final normalization steps results in a so called \emph{improved Fisher Kernel} and is -- in combination with a linear SVM \cite{cortes1995support} -- equivalent to the transformation of the raw FV using the Hellinger's kernel function \cite{perronnin2010improving}.

\section{Explaining Classification Decisions}
\label{sec:decomposing}
Most predictors, including linear SVMs over Fisher vectors, incorporate several layers of non-linear mappings, resulting in a non-linear black box with respect to the dependency of the prediction on its pixel inputs. In this section we introduce the concept of Layer-wise Relevance Propagation (LRP) \cite{BacBinMonKlaMueSam15} as a way to compute partial prediction contributions -- or \emph{relevance values} $R$ -- for intermediate and input representations based on the final classifier output. LRP acts on a single test-image similar to the work in \cite{DBLP:conf/eccv/ZeilerF14} and to partial-derivative based methods such as \cite{DBLP:journals/corr/SimonyanVZ13}. We refer the reader to \cite{DBLP:journals/corr/SamekBMBM15} for a comparison of these three explanation approaches.

\subsection{Layer-wise Relevance Propagation}
Layer-wise Relevance Propagation decomposes the mappings performed during prediction time to attribute to each component of the input its \emph{share} with which it contributes to the classifier output, explaining its relevance to the prediction output in its given state. This unsupervised process of decomposition is in principle applicable to any kind of model, resulting in high (positive) output values $R$ identifying properties of the input speaking for the presence of the prediction target and low (or even negative) scores indicating no or negative contribution. The conservation principle inherent to LRP ensures that no amount of relevance is gained or lost in between layers of computation,
\begin{align}
\sum\limits_{i} R_i^{(k)} = \sum\limits_{j} R_j^{(k+1)}
\label{eq:conservation}
\end{align}
where $R_i^{(k)}$ signifies the relevance value attributed to the $i$th computation unit or dimension at the $k$th computation layer of the prediction pipeline, and where the sums run over all units of the corresponding layers. In the context of an image classification problem, iterating LRP from the classifier output to the input layer results in outputs $R^{(1)}_p$ for each pixel $p$, with
\begin{align}
f(\x) = \sum_p R^{(1)}_p
\label{eq:conservation2}
\end{align}
and $f(\x)$ being equal the output layer relevance values.
In \cite{BacBinMonKlaMueSam15} examples have been given for decompositions of neural network architectures and Bag of Words feature extraction pipelines satisfying the above constraints. 

LRP propagates the relevance $R$ back from the output of a mapping towards its inputs. In a neural networks, a neuron maps a set of inputs $\{x_i\}$ to an output $x_j$ with monotonously increasing activation function $g(\cdot)$
\begin{align*}
x_j & = g(z_j) \quad \mathrm{with} \quad z_j = \sum_{i} z_{ij} ~,~ z_{ij}= w_{ij}x_i
\end{align*}
where the sum runs over all input neurons contributing to the activation of neuron $x_j$. The goal is to compute a relevance $R_i$ for input $x_i$ when relevances $R_j$ for outputs $x_j$ are given. \cite{BacBinMonKlaMueSam15} has introduced two possible formulas for relevance propagation
\begin{align}
R_i & = \sum_{j: i \rightarrow j} \frac{z_{ij}}{ z_j + \eps\cdot \mathrm{sign}( z_j )  } R_j \label{eq:lrp1}\\
R_i & = \sum_{j: i \rightarrow j} \Big( \alpha\frac{z_{ij}^{+}}{ z_j^{+}  } -\beta \frac{z_{ij}^{-}}{ z_j^{-}  } \Big) R_j,\label{eq:lrp2}
\end{align}
where $\sum_{j: i \rightarrow j}$ denotes a sum of all mappings which take $x_i$ as input. $z_{ij}^{+}$ denotes the positive part of the term, i.e.~$\max(0,z_{ij})$, $z_j^{+}$ is the sum over these positive parts. $z_{ij}^{-}$ is defined analogously as the negative part. The same paper has introduced a method to compute relevances for Bag of words (BoW) vectors, however, it tacitly assumed that BoW mappings are dominantly non-negative. For Fisher vectors this assumption does not hold, as the features are derivatives with respect to parameters. For this reason we propose a modified approach.

\subsection{LRP for Fisher Vector Classifiers}
\label{sec:decomposingFV}
Our variant to use LRP for Fisher vectors starts with writing the linear SVM as a mapping of features
\begin{align*}
f(x) = b+ \sum_{i} \alpha_i y_i \sum_{d=1}^D\phi(x_i)_d \phi(x)_d,
\end{align*}
where $x$ is a raw Fisher vector, and $\phi(x)$ realizes its normalization. In consistency with the first LRP formula, we define $R^{(3)}(x)$ as
\begin{align}
R^{(3)}_d &=  \sum_{i} \alpha_i y_i \phi(x_i)_d \phi(x)_d + \frac{b}{D}
\end{align}
From here on we apply for the mapping of local features $l$ to Fisher vectors $x$, equation \eqref{eq:lrp1} instead of the approach used in \cite{BacBinMonKlaMueSam15}. We can write the d-th dimension of the Fisher vector $x_d=\sum_l m_d(l)$. This is a mapping of local features $l$ onto the Fisher vector as a set of outputs $(x_d)_{d=1}^D$. We apply equation \eqref{eq:lrp1} with $z_{ld}= m_d(l)$. $m_d(l)$ is given in the notation of Section \ref{sec:fisher} as the term from equation \ref{eq:fvassemble}: 
\begin{align}
&&\hspace*{-1.3cm}m_{(d)}(l)=\Psi_\lambda(l)_{(d)}
\label{eq:fv_mmap}
\end{align}
Pixel-wise relevance scores $R^{(1)}_p$ are then computed by uniformly distributing for all local features $l$ the relevance scores $R_{l}^{(2)}$ onto the set of pixels $p$ covered by the receptive field of $l$, resulting in a \emph{heatmap} which can be visualized.
The decomposition process with explicit redistribution formulas is depicted in Figure \ref{fig:fv}.

\subsection{Measuring Context with LRP}
The distribution of positive relevance mass in a heatmap can be used for assessing the importance of context for a particular image classification task. If bounding box annotation are available (as for the Pascal VOC dataset), we can compute the \emph{outside-inside relevance ratio} metric defined as:
\begin{align}
\mu &=
\frac{
\frac{1}{|P_{\textrm{out}}|} \sum\limits_{q \in P_{\textrm{out}}} R_q^{(1)}
}
{
\frac{1}{|P_{\textrm{in}}|}\sum\limits_{p \in P_{\textrm{in}}} R_p^{(1)}
}
\label{eq:bbratio}
\end{align}
with $|\cdot |$ being the cardinality operator and $P_{\textrm{out}}$ and $P_{\textrm{in}}$ being the set of pixels outside and inside the bounding box, respectively. A high relevance ratio indicates that the classifier uses a lot of context to support the decision. A low relevance ratio indicates that the classifier focuses instead on the object to support its decision. Note that this measure can not be to $100\%$ accurate in most cases, since for example the bounding box areas of slim but obliquely angled objects, for example, aeroplanes photographed during lift-off, will also cover a considerable amount of image background.

\section{Experimental Evaluation}
\label{sec:experiments}

\subsection{Basic Setup}
All measurements are carried out on PASCAL VOC2007 \cite{DBLP:journals/ijcv/EveringhamEGWWZ15} test data. Fisher vectors are computed using the encoding evaluation toolkit (version 1.1) from \cite{chatfield2011devil} with settings as in this paper. The Fisher vectors are trained on the trainval part of the same dataset. The neural network is finetuned on the trainval part of PASCAL VOC2012, starting from the BVLC reference classifier of the Caffe package \cite{jia2014caffe} with a base learning rate of $0.001$ using a multi-label hinge loss. As we are interested in the ability of a neural net to use context, we do not use the bounding box ground truth to extract image patches which cover parts of bounding boxes. Instead we create 4 edges and one center crop per image together with mirroring, resulting in 10 training patches per image. Test scoring is done in the same fashion. This corresponds to a setting with only a few number of test windows, in which one would use larger patches during training and testing. The region-wise scores are computed for FV as described in Section \ref{sec:decomposing} using equation \eqref{eq:lrp1} with parameter $\eps=1$ and $\eps=100$. For neural nets we used equation \eqref{eq:lrp1} with $\eps=1$,$\eps=100$ and equation \eqref{eq:lrp2} with $\beta=1,\alpha=2$. Random perturbations for Fisher vectors were achieved by randomly sampling local features from the GMM.

\begin{table*}
\small
\centering
\begin{tabular}{|c||c|c|c|c|c|c|c|}
\hline
&    { \bf aeroplane} & { \bf bicycle} & { \bf bird} & { \bf boat} & { \bf bottle} & { \bf bus} & { \bf car}\\ 
{\bf Fisher} & 79.08\% & 66.44\% & 45.90\% & 70.88\% & 27.64\% & 69.67\% & 80.96\% \\ 
{\bf DeepNet} & 88.08\% & 79.69\% & 80.77\% & 77.20\% & 35.48\% & 72.71\% & 86.30\% \\\hline 
&    { \bf cat} & { \bf chair} & { \bf cow} & { \bf diningtable} & { \bf dog} & { \bf horse} & { \bf motorbike}\\ 
{\bf Fisher} & 59.92\% & 51.92\% & 47.60\% & 58.06\% & 42.28\% & 80.45\% & 69.34\% \\ 
{\bf DeepNet} & 81.10\% & 51.04\% & 61.10\% & 64.62\% & 76.17\% & 81.60\% & 79.33\% \\ \hline 
&    { \bf person} & { \bf pottedplant} & { \bf sheep} & { \bf sofa} & { \bf train} & { \bf tvmonitor} & { \bf mAP}\\ 
{\bf Fisher} & 85.10\% & 28.62\% & 49.58\% & 49.31\% & 82.71\% & 54.33\% & 59.99\% \\ 
{\bf DeepNet} & 92.43\% & 49.99\% & 74.04\% & 49.48\% & 87.07\% & 67.08\% & 72.12\% \\ \hline
\end{tabular}
\vspace{10pt}
\caption{Prediction performance of the trained Fisher model and deep network in average precision (AP) per class.}
\label{tab:cnnMAP}
\end{table*}


\begin{figure*}
\centering
\parbox{0.12\textwidth}{\centering Image}%
\parbox{0.12\textwidth}{\centering Heatmap\\(bike)}%
\parbox{0.12\textwidth}{\centering Image}%
\parbox{0.12\textwidth}{\centering Heatmap\\(person)}%
\parbox{0.12\textwidth}{\centering Image}%
\parbox{0.12\textwidth}{\centering Heatmap\\(cat)}%
\parbox{0.12\textwidth}{\centering Image}%
\parbox{0.12\textwidth}{\centering Heatmap\\(person)}\\
\includegraphics[width=0.24\textwidth]{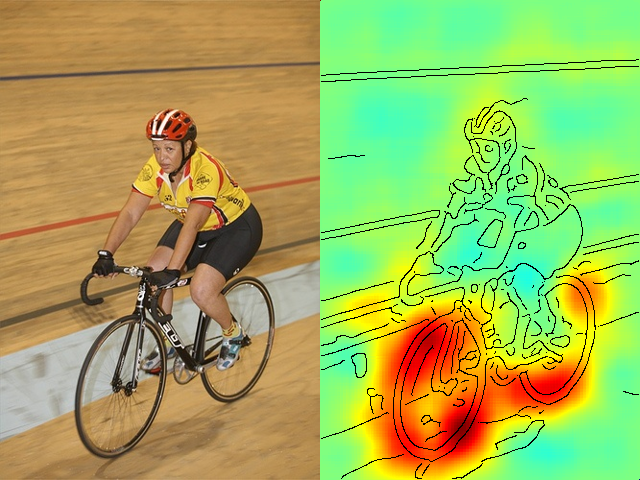}%
\includegraphics[width=0.24\textwidth]{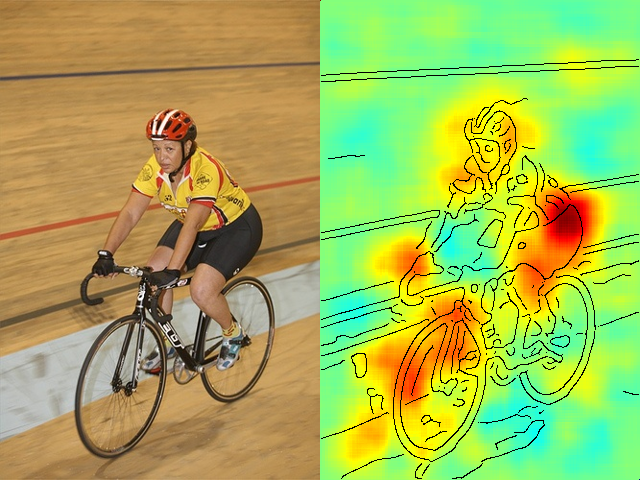}%
\includegraphics[width=0.24\textwidth]{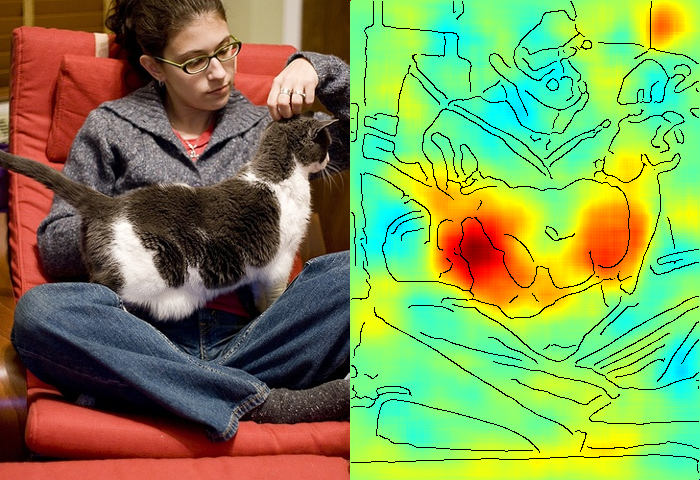}%
\includegraphics[width=0.24\textwidth]{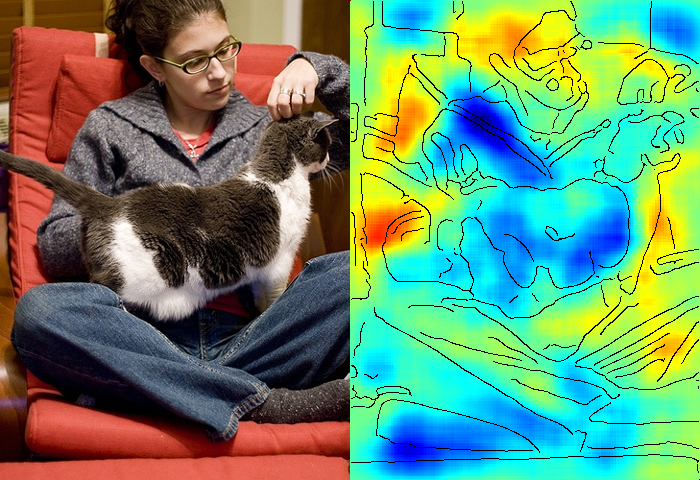}\\[2mm]
\parbox{0.165\textwidth}{\centering Image}%
\parbox{0.165\textwidth}{\centering Heatmap\\(train)}%
\parbox{0.165\textwidth}{\centering Image}%
\parbox{0.165\textwidth}{\centering Heatmap\\(train)}%
\parbox{0.165\textwidth}{\centering Image}%
\parbox{0.165\textwidth}{\centering Heatmap\\(dining table)}\\
\includegraphics[width=0.33\textwidth]{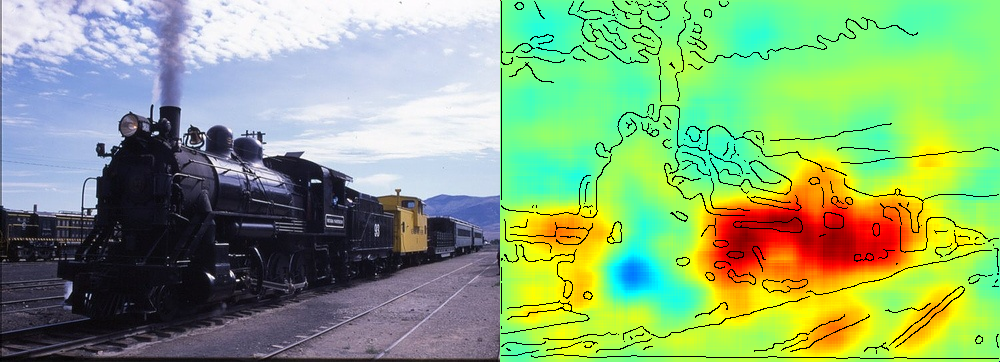}%
\includegraphics[width=0.33\textwidth]{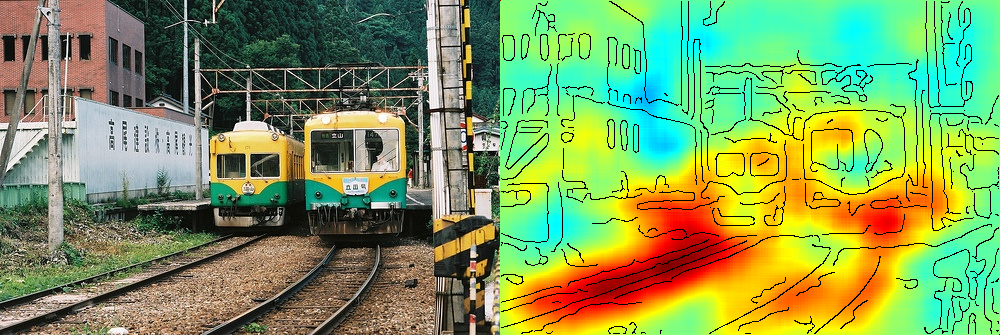}%
\includegraphics[width=0.3\textwidth]{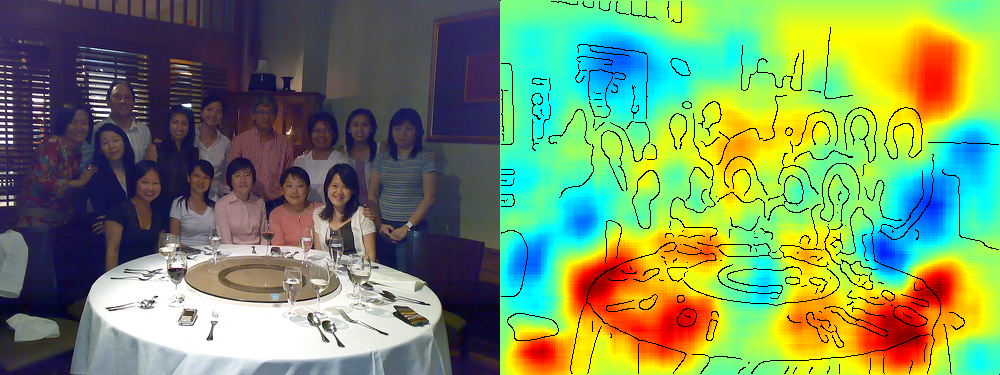}%
\caption{Images shown next to the heatmaps computed by application of LRP on the FV model when considering the prediction score for a particular class.}
\label{figure:heatmaps}
\end{figure*}

\begin{figure*}[th] 
\centering
\includegraphics[width=0.32\textwidth]{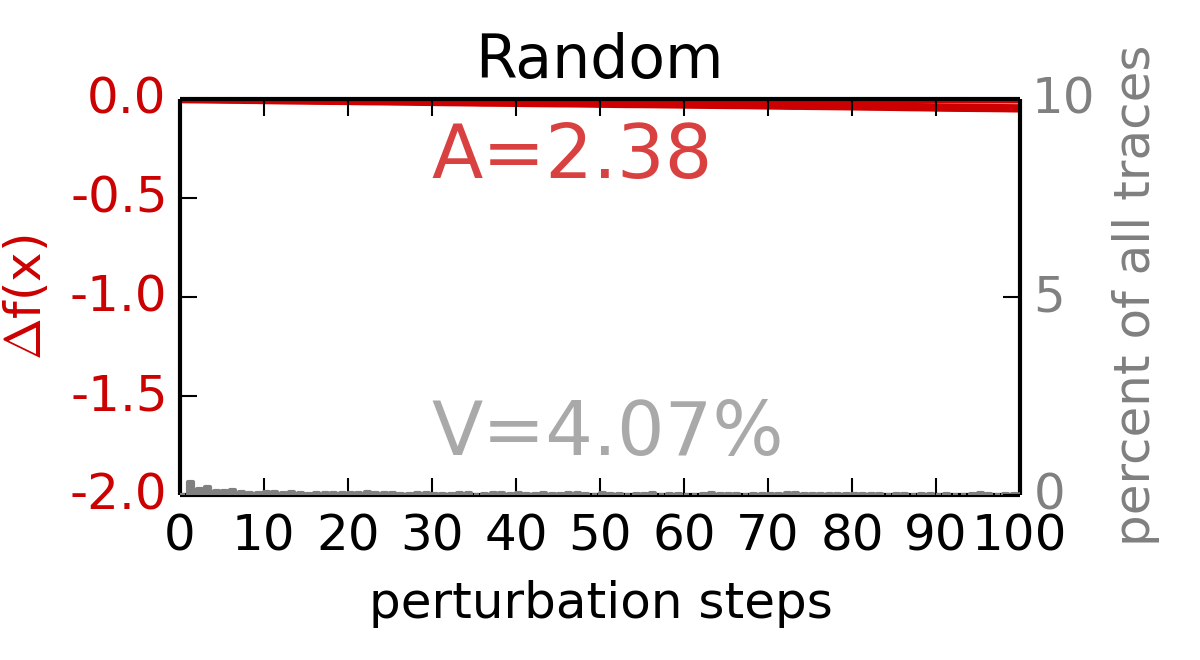}
\includegraphics[width=0.32\textwidth]{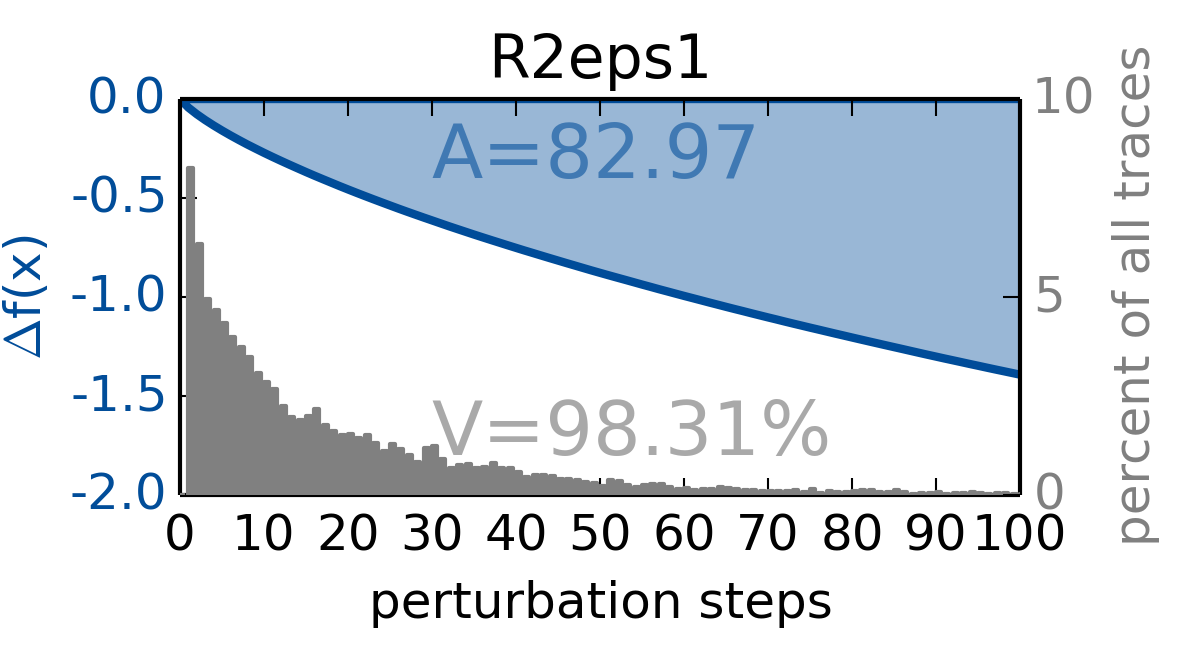}
\includegraphics[width=0.32\textwidth]{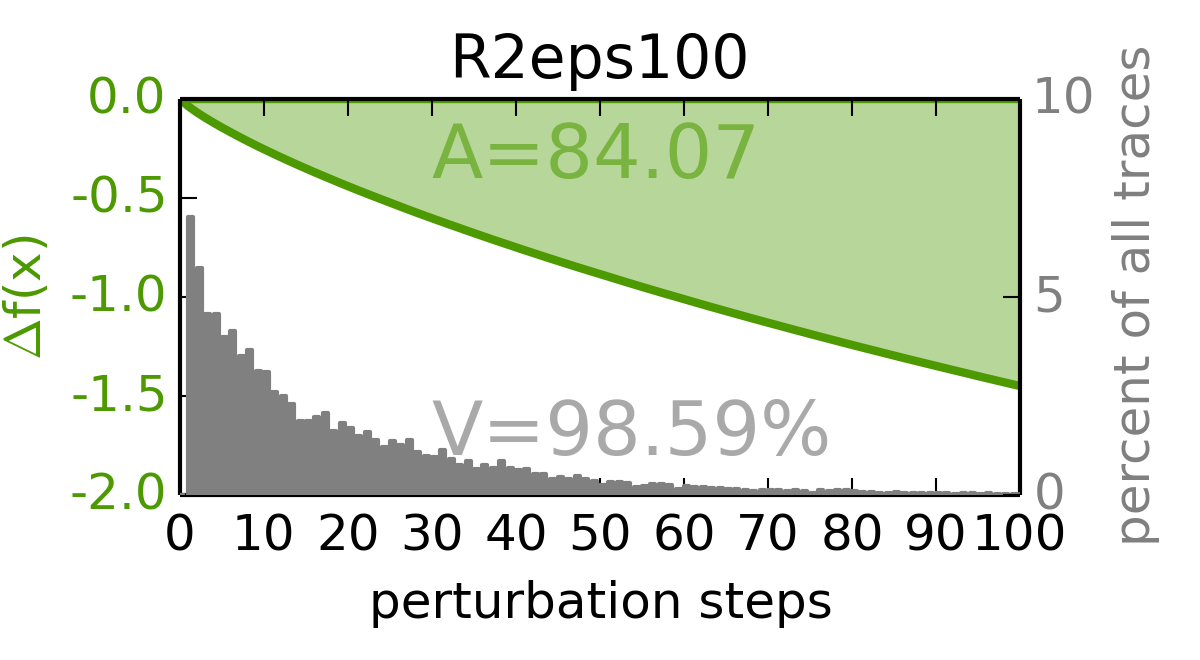}
\caption{Heatmap Quality Measurements for Fisher Vectors. The value A measures the area above the curve between the original prediction $f(\x)$ and the averaged perturbed prediction at step $i$ in the sequence of regions. $f(\x) - f(\x^{(i)}_{\MoRF})$. V represents the fraction of all perturbation sequences for which the prediction switched sign at some step in the sequence, with the gray bar chart showing how many sample traces changed class at each point of measurement.}
\label{fig:lfreplace}
\end{figure*}

\subsection{Are Fisher Explanations Meaningful ?}
The first step before measuring the amount of context is to validate whether the computed scores for a pixel or a region are meaningful at all. Figure \ref{figure:heatmaps} depicts heatmaps computed on exemplary test images of the Pascal VOC data set considering the prediction score for a particular class. The quality of these explanations can be intuitively assessed by a human, e.g., it makes perfectly sense that the Fisher vector classifier finds that wheels are relevant for the class `bike', rail tracks are indicative for the class `train' and tableware is important for classifying images of class `dining table'. These examples show that the largest part of the relevance mass does not necessarily need to lie on the object, on the contrary it may be the context which is the informative part. 

In order to objectively validate that the Fisher vector heatmaps are meaningful we evaluate the decrease of the prediction score under perturbations. The idea is that a region such as an image patch is highly relevant, if modifying it results for most modifications in a sharp decline of the prediction for the whole image. Modifying a region is done by randomly perturbing the pixels with noise. The prediction score is averaged over a number of random perturbations, in order to capture the average change of the classifier. 

This notion of relevant regions can be used for evaluation of region scores by sorting image regions along descending scores. Then, for each region in the sequence the average decrease of predictions is measured. The result is a graph as a function of the sequence index. Thus under this evaluation scheme, a region-wise score performs well if it assigns highest scores to regions which are most sensitive on average under perturbations and yield the sharpest decline of the prediction score. \cite{DBLP:journals/corr/SamekBMBM15} introduced this setup and evaluated the methods of \cite{DBLP:conf/eccv/ZeilerF14,DBLP:journals/corr/SimonyanVZ13,BacBinMonKlaMueSam15} for deep neural networks tested on ImageNet \cite{ILSVRC15}, SUN397 \cite{xiao2010sun} and MIT Places \cite{zhou2014learning}. Here we show that LRP scores computed are also meaningful for Fisher vectors. Figure \ref{fig:lfreplace} shows this comparison against random orderings for scores computed. The LRP scores produce a more meaningful ordering than random sequences which motivates its use to define a measure for context. 


\begin{figure*}
\centering
\parbox{0.33\textwidth}{\centering Image}%
\parbox{0.33\textwidth}{\centering Fisher}%
\parbox{0.33\textwidth}{\centering DeepNet}\\
\includegraphics[width=0.66\textwidth]{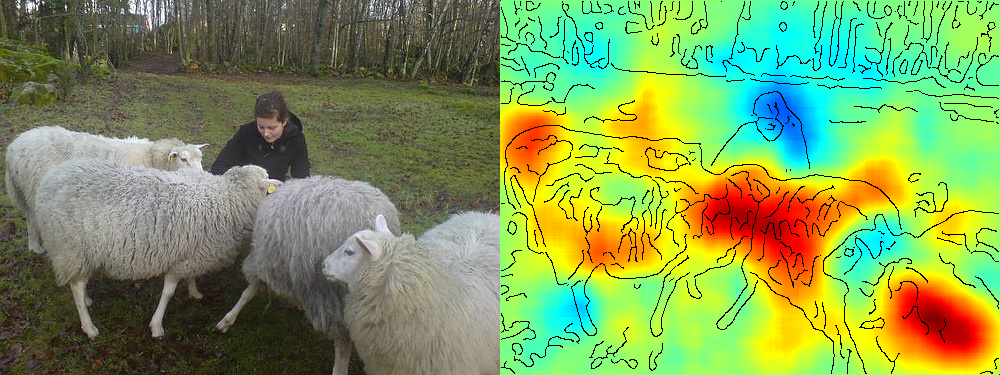}%
\includegraphics[width=0.33\textwidth,trim=227 28.5 0 28.5,clip=true]{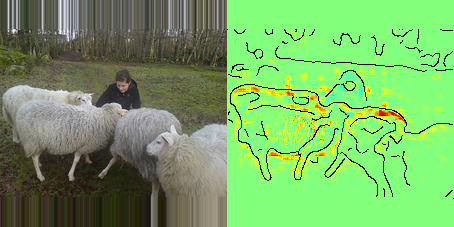}\\[1mm]
\includegraphics[width=0.66\textwidth]{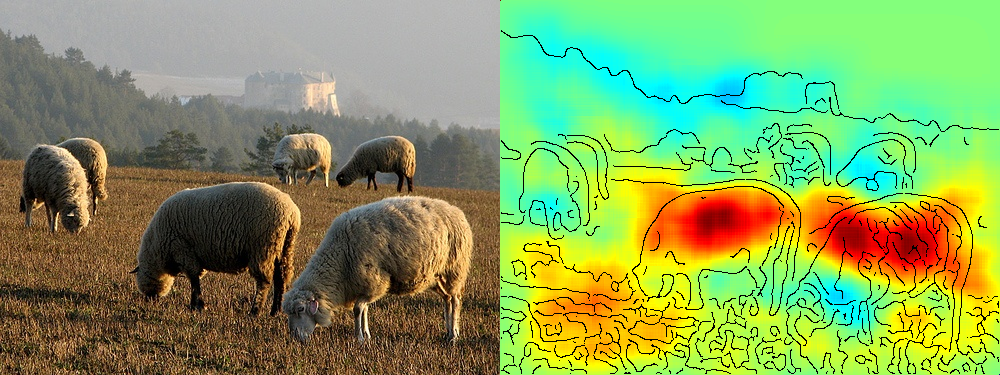}%
\includegraphics[width=0.33\textwidth,trim=227 28.5 0 28.5,clip=true]{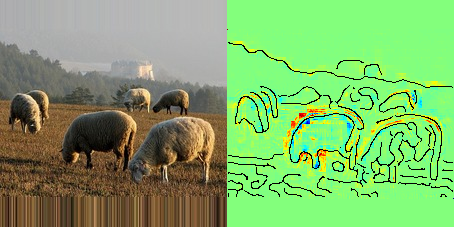}
\caption{Images of the class ``sheep'', processed by the FV and DNN models and heatmapped using LRP.}
\label{figure:sheeps}
\end{figure*}

\subsection{Shallow vs. Deep Features}
\label{section:shallowdeep}
We investigate in the light of the LRP framework what are the differences of strategies used to classify images between (1) a shallow model operating on high-resolution images: the FV model, and (2) a deep model operating on lower-resolution images: the DNN model. We consider first the class ``sheep'' for which the DNN produces much better predictions than the FV model (25\% superior accuracy in absolute terms according to Table \ref{tab:cnnMAP}).

Example of two images of class ``sheep'' and the corresponding heatmaps for the FV and DNN models are shown in Figure \ref{figure:sheeps}. The LRP analysis reveals that the FV and DNN models use clearly different strategies to predict the class:

The FV model bases its decision on the wool texture typical of the sheep and available at high-resolution, but ignores the exact shape of the sheep. Interestingly, relevance is also allocated to the context (here, positive relevance for the grass and negative relevance for the human face), indicating that the context is an essential component of the classifier and modulates the prediction score positively or negatively.

On the other hand, the DNN assigns a large proportion of heat to the border of the sheep, thus, showing that the shape of the sheep (e.g. its contour) is exploited in order to improve the prediction. Furthermore, for the DNN, the LRP method does not assign relevance to contextual elements such as the grass, or the human face, nor to the wool texture of the sheep, which is harder to detect due to the low resolution of images given to the CNN.

Overall, the LRP analysis indicates that the far superior predicting power of the DNN model must be attributed in largest part to the ability to model the exact shape of the sheep, making all remaining contextual or texture features less relevant. On the other hand, the less accurate FV model does benefit from the weak correlations between object class, texture and context to improve prediction quality.

\begin{figure*}
\begin{minipage}[t]{0.465\textwidth}
\parbox{0.33\textwidth}{\centering Image}%
\parbox{0.33\textwidth}{\centering FV}%
\parbox{0.33\textwidth}{\centering DNN}\\
\includegraphics[width=0.66\textwidth]{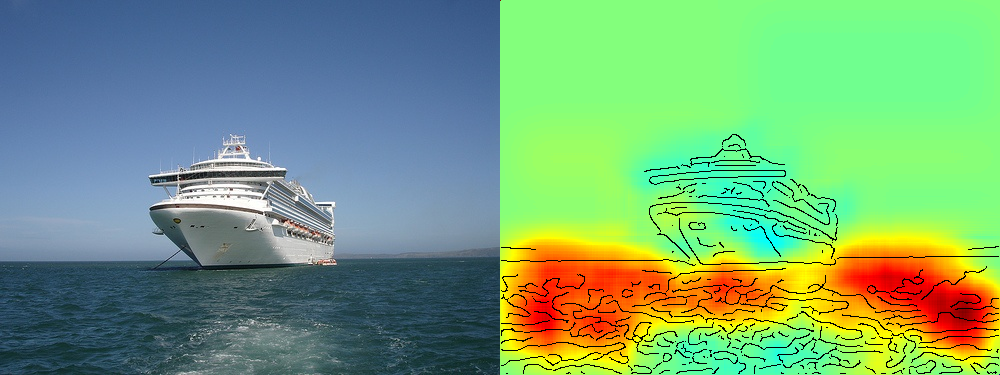}%
\includegraphics[width=0.33\textwidth,trim=227 28.5 0 28.5,clip=true]{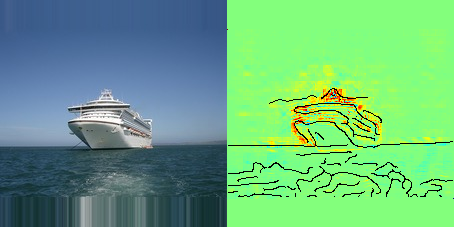}\\[1mm]
\includegraphics[width=0.66\textwidth]{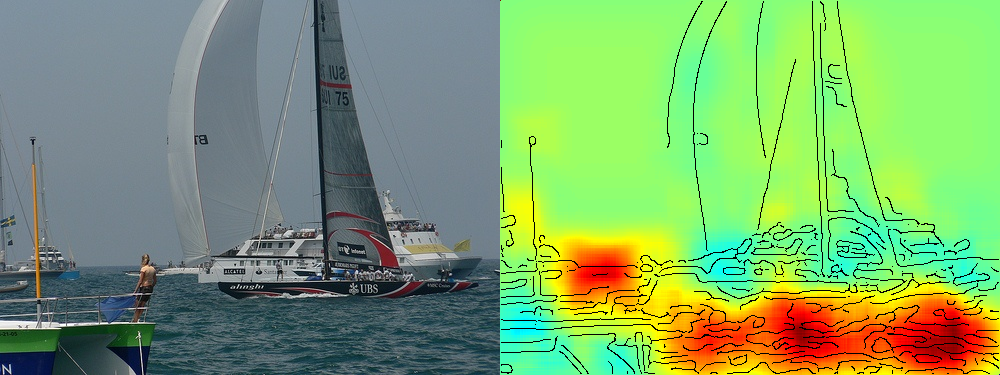}%
\includegraphics[width=0.33\textwidth,trim=227 28.5 0 28.5,clip=true]{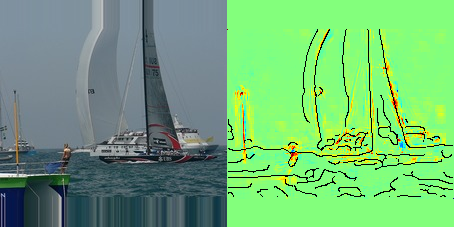}\\[1mm]
\parbox[t]{0.33\textwidth}{\centering\vspace*{-1cm} average\\ heatmaps}%
\includegraphics[width=0.33\textwidth,trim=0 0 0 0,clip=true]{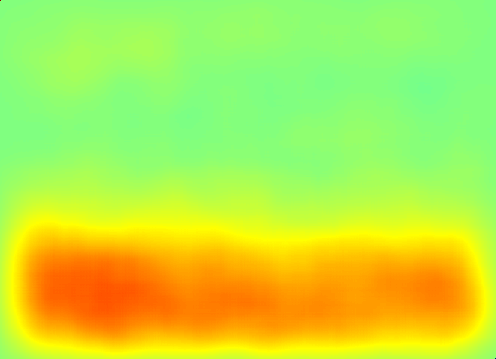}%
\includegraphics[width=0.33\textwidth,trim=0 0 0 0,clip=true]{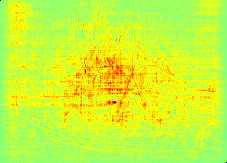}
\end{minipage}\hfill
\begin{minipage}[t]{0.525\textwidth}
\raggedleft
\parbox{0.33\textwidth}{\centering Image}%
\parbox{0.33\textwidth}{\centering FV}%
\parbox{0.33\textwidth}{\centering DNN}\\
\includegraphics[width=0.66\textwidth]{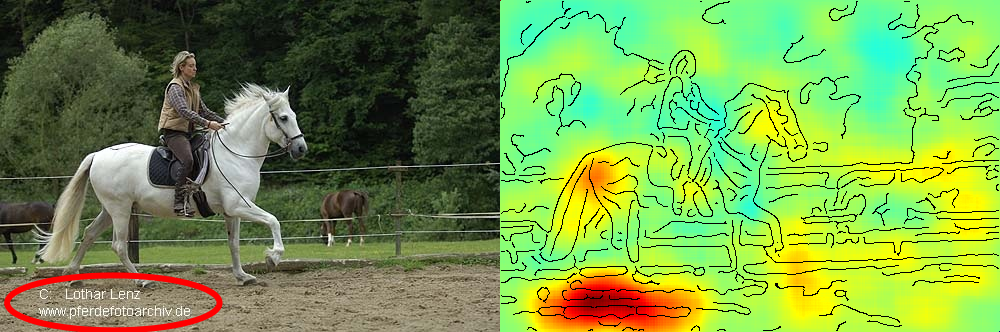}%
\includegraphics[width=0.33\textwidth,trim=227 38.5 0 38.5,clip=true]{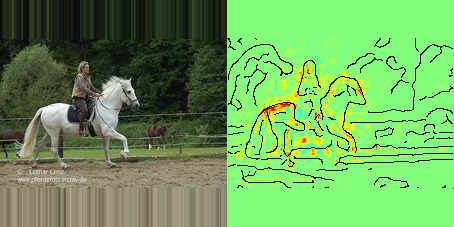}\\[1mm]
\includegraphics[width=0.66\textwidth]{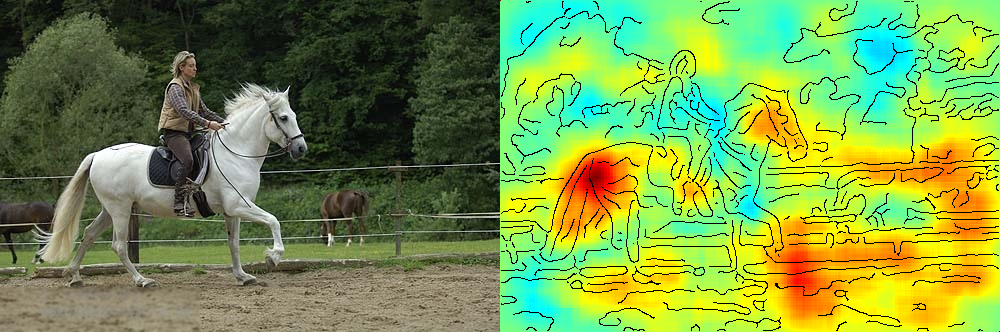}%
\includegraphics[width=0.33\textwidth,trim=227 38.5 0 38.5,clip=true]{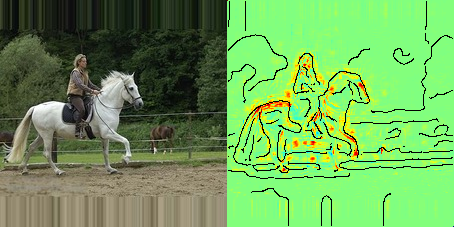}%
\\[1mm]
\parbox[t]{0.33\textwidth}{\centering\vspace*{-1cm} average\\ heatmaps}%
\includegraphics[width=0.33\textwidth,trim=0 0 0 0,clip=true]{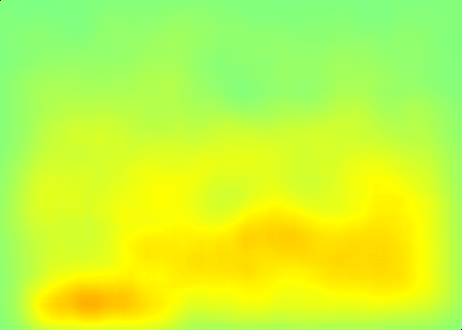}%
\includegraphics[width=0.33\textwidth,trim=0 0 0 0,clip=true]{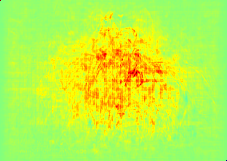}
\end{minipage}%
\vskip 3mm
\caption{Top: Images of the classes ``boat'' and ``horse'', processed by the FV and DNN models and heatmapped using LRP. Bottom: Average heatmap scores over a random sample (of size between 47 and 177) of the distribution for each class and model. On the second image of class ``horse'', the copyright tag (marked by the red ellipse) has been removed.}
\label{figure:boatshorses}
\end{figure*}

\subsection{Test Error and Model Quality}

For other classes, it can be observed in Table \ref{tab:cnnMAP} that test error of the FV model is almost on par with the one of the DNN. We investigate whether high test accuracy is predictive of the ability of the model to extract meaningful features for a given class, or whether the decision is based mostly on undesirable contextual or artefactual features.

\paragraph{Contextual features}

As an illustrative example, we consider the class ``boat'', where the performance of the DNN superior by less than 7\% in absolute terms to the FV model. (Note that for other classes such as ``sheep'' or ``bird'', the DNN performance is superior by 25\% or more.) It is tempting to conclude that, for the class ``boat'', both models should have learned a set of features of similarly high quality.

LRP analysis gives a different answer: Figure \ref{figure:boatshorses} (left) shows the heatmaps produced by the FV and DNN models on two archetypical images of the class ``boat''. For the DNN, LRP assigns most of the relevance to pixels corresponding to the actual boat. On the other hand, for the FV model, LRP assigns most relevance to the water below the boat (i.e. the FV model does not recognize the object itself, but its context). The heat distribution of average heatmaps (computed over all landscape-format images of the class ``boat'') corroborates what was observed for two selected images, in particular, a focus of the FV model on the bottom part of the image where water usually is, and a focus of the DNN model on the middle part of the image where the boat typically is.

We can conclude from the LRP analysis, that while both classifiers have a roughly similar level of accuracy on the test images with class ``boat'', FV's performance is likely to decrease drastically if one were to consider boats located outside the water as test images. On the other hand, performance of the DNN would be less affected. Therefore, test error is a superficial predictor of model quality in this case.

\paragraph{Artefactual features}

A second example where high accuracy does not necessarily translates into high quality features is for the class ``horse''. This class is predicted with similar accuracy by the FV and DNN models (approximately 1\% difference in accuracy).

Figure \ref{figure:boatshorses} (right) shows a LRP heatmaps for the FV and DNN model on an image of horse. While the DNN assigns relevance on the actual object ``horse'', the FV assigns almost all relevance in the bottom-left corner the image, where careful inspection of the image reveals the presence of a copyright tag. Thus, the decision of the FV model is in large part based on the presence of the copyright tag, which is discriminative of the class horse. Removing the copyright tag completely changes the FV heatmap, but does not change significantly the DNN heatmap.

If the copyright tag is removed, the DNN is still able to predict the image because the pixels that support its decision are not affected. On the other hand, FV model prediction quality will be considerably reduced. The systematic focus of the FV model on the copyright tag is confirmed in the average heatmap, where the bottom-left corner is assigned large amount of heat. Therefore, for this class again, test error does not predict well model quality.

\begin{figure*}
\includegraphics[width=1\textwidth]{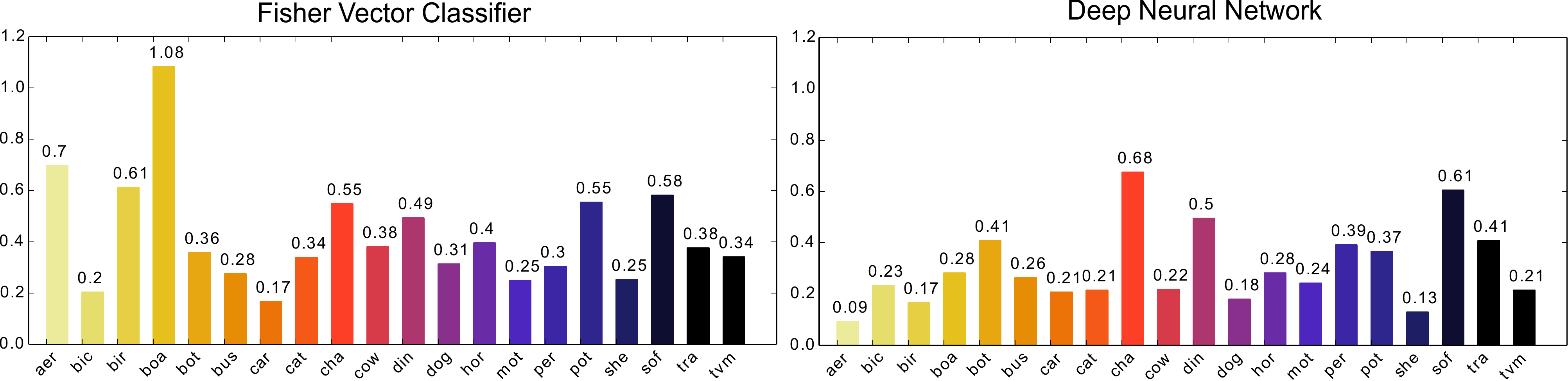}%
\caption{Outside-inside relevance ratio as computed by equation \ref{eq:bbratio} for the 20 classes of the Pascal VOC 2007 dataset. Left: ratios for the FV model. Right: ratios for the DNN model.}
\label{figure:context}
\end{figure*}

\subsection{Quantitative analysis of context use}

While we have so far provided a qualitative interpretation of FV heatmaps for examples and classes of interest, we can more systematically measure whether the model uses context or the actual object, by measuring for each classes and models the outside-inside relevance ratio $\mu$ computed by equation \ref{eq:bbratio}.

Results are shown in Figure \ref{figure:context}. Generally, the FV model uses more context than the DNN, as evidenced by a higher relevance ratio. However, there are significant differences between classes:

Classes where the use of context by the FV model is particularly high are ``boat'' and ``airplane'', the first of which we have studied qualitatively in the previous section. For these two respective classes, the water and the sky are important contextual elements that support the decision of the Fisher model, due to their strong correlation. Another group of classes with high context of the Fisher model are ``chair'', ``diningtable'', ``pottedplant'' and ``sofa'' which share a semantic of indoor room sceneries.

For other classes such as ``bicycle'', ``car'', ``motorbike'', or ``sheep'', the Fisher model does not use much context. For the first three classes, the urban environment surrounding these classes is not predictive of the object being detected (i.e. it could not discriminate between these three classes based on the context only). For the last class, as it has been discussed in Section \ref{section:shallowdeep}, the wool texture of the sheep (which lies inside the sheep bounding box) is a reasonable predictor for the class ``sheep'', although the actual object sheep (i.e. defined by its shape or contour) is not being used.

As for deep neural networks, classes with least context usage are ``aeroplane'', ``bird'', ``sheep'', ``dog'', ``car'', ``cat'' and ``tvmonitor''. Each of those is associated with a significantly better score achieved by the DNN.

\section{Conclusion}
\label{sec:discussion}

In this paper, we have analyzed what make Fisher vector models (FV) and deep neural networks (DNN) decide for a particular class. To achieve this, we have employed a heatmapping technique that determines what pixels in the image are used by a classifier to support its decision. The technique called layer-wise relevance propagation and originally developed for neural networks \cite{BacBinMonKlaMueSam15} was extended to Fisher vector models, and validated using the method by \cite{DBLP:journals/corr/SamekBMBM15}.

Our novel comparative analysis of FV and DNN classifiers corroborates empirically previous intuition relating the architecture of the classifier to the features it is able to extract. In particular, our analysis shows that the FV model compensates its lack of depth by the use of contextual information---potentially artefacts---that are weakly correlated to the object class. We thus demonstrate that the generalization capability of Fisher vector models can be overstated if test images also include similar context.

On the other hand, DNNs base their decision on the actual object to detect and ignores its context. This focus on object detection has to be attributed to the higher overall predictive accuracy of the model, that removes the need for contextual information---even if the latter is discriminative. The focus on detection must also be attributed to the deep multitask properties of the DNN that favors composition of natural image features over lower-level features such as copyright text.

These results argue in favor of incorporating heatmapping techniques into the data collection and model selection processes. The interpretable visual feedback that heatmaps provide can be used in particular to verify that the considered classifier bases its decision on the right set of features, and in the contrary case, select another model, or extend the dataset in a way that artefactual features can no longer support the classification decision.

{\small
\bibliographystyle{ieee}
\bibliography{paperArxiv}
}

\onecolumn
\appendix

\section{Details on the Computation of Fisher Vector Embeddings}
The representation of an image data set as FVs starts with the computation of a Gaussian Mixture Model (GMM) as a soft vocabulary of visual prototypes. The $K$ components $\lambda = \lbrace (\pi_k, \mu_k, \Sigma_k) \rbrace_{k=1..K}$~ of the GMM are fitted onto a set of local descriptors $l$ extracted from the training images, where $\mu_k$ is the mean vector of the $k$th mixture component and $\Sigma_k$ its covariance matrix which is assumed to be diagonal, e.g. $diag\left(\Sigma_k\right) = \sigma_k$ and all off-diagonal entries are $0$.
The parameter $\pi_k$ is the mixture weight of component $k$, with $\sum_k \pi_k = 1$ and $\forall k : \pi_k \geq 0$.
A \emph{full} FV descriptor of an image measures the average $0$th (soft mapping weight), $1$st (deviation from mean) and $2$nd (variance) moment of all local descriptors sampled from the image area \cite{sanchez2013image}. 
\begin{align}
\Psi_{\pi_k}(\l) & = \frac{1}{\sqrt{\pi_k}}\left( \gamma_k(\l) - \pi_k \right) \nonumber\\
\Psi_{\mu_k}(\l) & = \frac{1}{\sqrt{\pi_k}}\gamma_k(\l)\left( \frac{\l - \mu_k}{\sigma_k} \right) \nonumber\\
\Psi_{\sigma_k}(\l)& = \frac{1}{\sqrt{\pi_k}}\gamma_k(\l)\frac{1}{\sqrt{2}}\left( \frac{\left(\l - \mu_k\right)^2}{\sigma_k^2} - 1 \right) \nonumber\\ 
\label{eq:fvmap}
\end{align}

with $\Psi_{\pi_k}(\l) \in \mathbb{R}$ ,  $\{~\l,~\Psi_{\mu_k}(\l),~\Psi_{\sigma_k}(\l)~\} \in \mathbb{R}^D$ and $\gamma_k(\l)$ returning the soft assignment of $l$ to the $k$th mixture component \cite{sanchez2013image}.
The FV embedding $\Psi_\lambda(\l)$ of a local descriptor $\l$ is achieved by concatenating the mapping outputs for all $K$ components into a $(1+2D)K$ dimensional vector, such that
\begin{align}
\Psi_\lambda(\l) = \left[\underbrace{\Psi_{\pi_1}(\l)~,~\dots~,\Psi_{\pi_K}(\l)}_{K~\textrm{ scalar entries}} , \underbrace{\Psi_{\mu_1}(\l),~\dots~,\Psi_{\mu_K}(\l)}_{K~\textrm{concatenated}~D\textrm{-dimensional vectors}} , \underbrace{\Psi_{\sigma_1}(\l),~\dots~,\Psi_{\sigma_K}(\l)}_{K~\textrm{concatenated}~D\textrm{-dimensional vectors}} \right]
\label{eq:fvassemble}
\end{align}

The final FV of an image is obtained by averaging over all $\Psi_\lambda(\l)$ resultung in the \emph{raw} FV representation\footnote{\emph{raw} FV in contrast to the \emph{improved} FV, as used in the main document to distinguish between the both.}, followed by power normalization to reduce the sparsity of the descriptor and and $\ell_2$-normalization to improve prediction performance \cite{perronnin2010improving}:
\begin{align}
\x ~&~ = \frac{1}{|L|} \sum\limits_{\l\in L} \Psi_\lambda(\l) ~&~ \textrm{mapping aggregation} \nonumber\\
\x ~&~ \leftarrow sign(\x)|\x|^\frac{1}{2} ~&~ \textrm{power normalization} \nonumber\\
\x ~&~ \leftarrow \frac{\x}{\|\x\|_2} ~&~ \ell_2\textrm{-normalization} \nonumber
\end{align}

\section{On the Equality of the Normalization Steps and the Hellinger's Kernel}
\label{sec:hellinger}

The work of \cite{sanchez2013image} references the equality of the application of the Hellinger's (Bhattacharyya) kernel function to a raw FV descriptor and the improved FV (power- and $\ell_2$-normalized FV) with a linear kernel function on top. Here, we explicitly show this equality. Assume a raw FV $\x$, the component-wise absolute of an input variable $|\cdot|$, and the $\ell_p$-norm of a vector $\|\cdot\|_p$. The improved FV defines itself as the application of power-normalization followed by $\ell_2$-normalization on top of a raw FV, e.g. interpreted as a mapping function $\Phi(\cdot)$ we receive
\begin{align}
\Phi(\x) &\ =\ \frac{sign(\x)|\x|^{\frac{1}{2}}}{\| sign(\x)|\x|^{\frac{1}{2}} \|_2}
\ =\  \frac{sign(\x)\sqrt{|\x|}}{\sqrt{\sum\limits_d \underbrace{sign(x_{(d)})^2}_{=1} | x_{(d)} |\underbrace{^{\frac{1}{2}\cdot 2}}_{=1}}}\\[+3px]
& =\ sign(\x)\sqrt{\frac{|\x|}{\sqrt{\sum\limits_d |x_{(d)}|}^2}}
\ =\ sign(\x)\sqrt{\frac{|\x|}{\|\x\|_1}} \nonumber
\end{align}
and with that
\begin{align}
k(\x,\y) = \sum\limits_d \Phi(\x)_{d}\Phi(\y)_{d} =\ \sum\limits_d sign(x_{(d)} y_{(d)})\underbrace{\sqrt{ \frac{|x_{(d)}|}{\|\x\|_1} \cdot \frac{|y_{(d)}|}{\|\y\|_1} }}_{\textrm{Hellinger's kernel}}
\end{align}
which enables us to compute component-wise relevance scores for each FV dimension $d$ by attributing the power normalization to the kernel function as
\begin{align}
R_d^{(3)} & = \sum\limits_{i}\alpha_i \Phi(\x_i)_{d}\Phi(\x)_{d} + \frac{b}{D}
\label{eq:R3_1}
\end{align}
In the case of an improved FV mapping (or the Hellinger's (Bhattacharyya) kernel function), the support vectors need not to be known explicitly, since $w = \sum_i \alpha_i \Phi(x_i)$, further simplifying Equation to
\begin{align}
R_d^{(3)} & = w_d\Phi(\x)_{d} + \frac{b}{D}
\label{eq:R3_2}
\end{align}

\section{Details on Relevance Decomposition for Fisher Vectors}

As already stated within the main document, the relevance decomposition for FV embeddings follows the definitions and constraints of \cite{BacBinMonKlaMueSam15}. As already stated in Section \ref{sec:hellinger}, relevance values $R_d^{(3)}$ for each dimension $d$ of the FV representation of an image can be easily computed as
\begin{align}
R_d^{(3)} & = \sum\limits_{i}\alpha_i \Phi(\x_i)_{d}\Phi(\x)_{d} + \frac{b}{D}
\label{eq:R3_1again}
\end{align}

The next step towards local explanations in pixel space is the computation of relevalce scores $R_{\l}^{(2)}$ for each local descroptor $\l$ contributing to the computation of the FV used for prediction. The work of \cite{BacBinMonKlaMueSam15} introduces the notion of a mapping function $m_{(d)}(\l)$ relating $\l$ to dimension $d$ in output space for Bag of Feature (BoF) models assuming positive mapping outputs exclusively. Reformulating $\Psi_{\lambda}(\l)$ in terms of $m_{(d)}(\l)$ to fit it into the LRP framework facilitates the computation of local feature relevance scores proportionally to their forward mapping contributions

\begin{eqnarray}
&&\hspace*{-1.3cm}m_{(d)}(\l) = 
\begin{cases} 
\frac{1}{\sqrt{\pi_k}}(\gamma_k(\l)-\pi_k) &;~d=k,~ k\in[1,K]\\
\frac{1}{\sqrt{\pi_k}}\gamma_k(\l)\left(\frac{l_{(r)}-\mu_{k,(r)}}{\sigma_{k,(r)}}\right)   &;~ d=K + D(k-1) + r, k\in[1,K], r\in[1,D]\\
\frac{1}{\sqrt{\pi_k}}\gamma_k(\l)\frac{1}{\sqrt{2}}\left(\frac{(l_{(r)}-\mu_{k,(r)})^2}{\sigma_{k,(r)}^2}-1\right)  &;~d=(1+D)K + D(k-1) + r , k\in[1,K], r\in[1,D]
\end{cases}
\label{eq:fv_mmap}
\end{eqnarray}

With that the proposed decomposition formula for BoF mappings  can directly be applied with only a minor adaption of $Z(\x)$ from Equation (26) in \cite{BacBinMonKlaMueSam15} to consider that $m_d(\l)$ may output mapping weights of both positive and negative signs for FVs :
\begin{eqnarray}
	R_{\l}^{(2)} = & \sum\limits_{d\not\in Z(\x)} R_d^{(3)} \frac{m_{(d)}(\l)}{\sum_{\l'\in L}m_{(d)}(\l')}\ +\ \xi
	\label{eq:R2decmop}
\end{eqnarray}
where
\begin{eqnarray}
Z(\x) = \left\{ d~|\forall \l : m_{(d)}(\l) = 0 \right\} \text{~and~} \xi = \sum\limits_{d\in Z(\x)} R_d^{(3)}\frac{1}{|L|} \nonumber
\end{eqnarray}

Pixel-wise relevance scores $R^{(1)}_p$ are then computed by uniformly distributing for all local features $\l$ the relevance scores $R_{\l}^{(2)}$ onto the set of pixels $p$ covered by the receptive field of $l$
\begin{align}
L(p) = & \lbrace \l | p \in \area(\l) \rbrace \nonumber\\
R_p^{(1)} = & \sum\limits_{\l \in L(p)} \frac{ R_{\l}^{(2)} }{ |\area(\l)| } 
\end{align}
resulting in a \emph{heatmap} which can be visualized.

Above approach to compute values $R_{\l}^{(2)}$ can in general -- when $m_{(d)}(\l)$ outputs values of both signs -- be expected to be numerically problematic when $\sum_{\l'} m_{(d)}(\l')$ becomes very small due to individual mappings $m_{(d)}(\l')$ cancelling each other out.
Divisions close to zero would then pronounce otherwise insignificant local descriptor weights for relevance distribution. This problem is known in the context of neural network relevance decompositions and has been discussed in \cite{BacBinMonKlaMueSam15}, which we will follow for extending Equation \ref{eq:R2decmop} to better satisfy the task at hand. Figure \ref{fig:lfreplace} will then compare how well the computed heatmap resulting from each decomposition method represents the classifier's perception according to the measurement procedure described in Section \ref{sec:alternatives_comparison}.

\begin{figure*}[th] 
\centering
\includegraphics[width=0.49\textwidth]{Random.png}
\includegraphics[width=0.49\textwidth]{R2eps1.png} \\
\includegraphics[width=0.49\textwidth]{R2eps100.png} 
\includegraphics[width=0.49\textwidth]{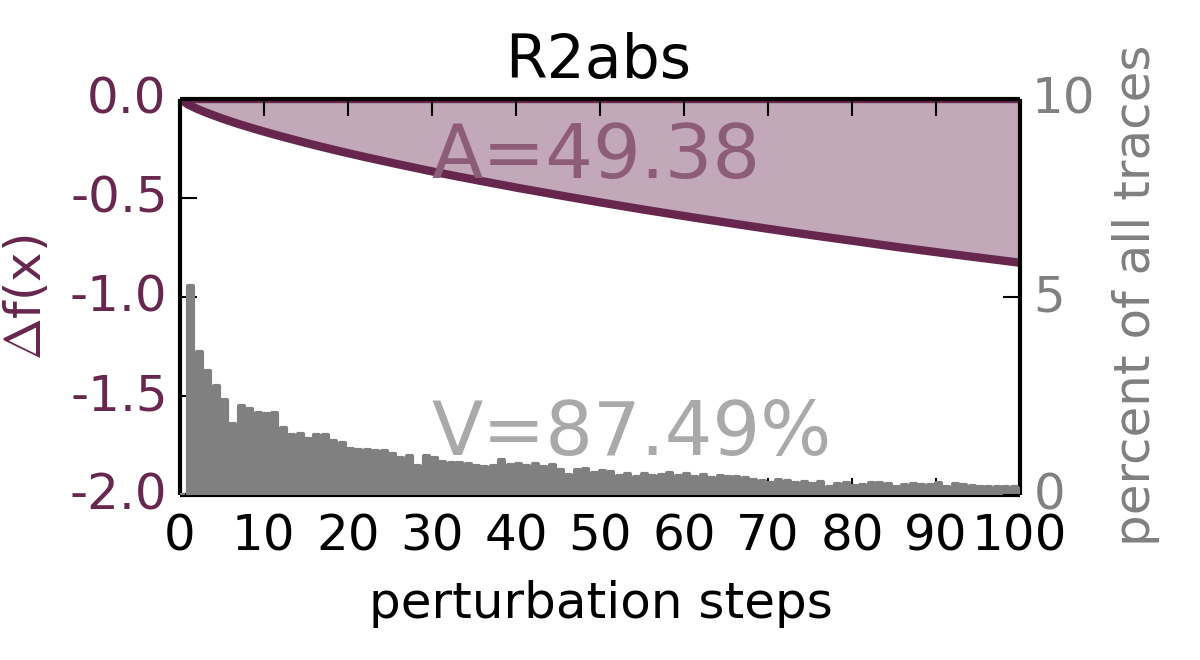}
\caption{Heatmap Quality Measurements for Fisher Vectors. The value A measures the area above the curve between the original prediction $f(\x)$ and the averaged perturbed prediction at step $i$ in the sequence of regions. $f(\x) - f(\x^{(i)}_{\MoRF})$. V represents the fraction of all perturbation sequences for which the prediction switched sign at some step in the sequence, with the gray bar chart showing how many sample traces changed class at each point of measurement. Despite violating the relevance conservation principle we have found the $\epsilon$-stabilized decomposition to produce the best heatmaps for the FV classifer. Therefore heatmaps computed with Equation \ref{eq:stable} and $\epsilon=100$ for all further evaluations.}
\label{fig:lfreplace}
\end{figure*}

\subsection{$\epsilon$-Stabilized Decomposition}
\label{sec:stable}

The first and probably most straight forward adaption of the decomposition formula introduced in \cite{BacBinMonKlaMueSam15} is the introduction of a numerical stabilizer $\epsilon>0$ to prevent (near) zero divisions
\begin{eqnarray}
	R_{\l}^{(2)} = & \sum\limits_{d\not\in Z(\x)} R_d^{(3)} \frac{m_{(d)}(\l)}{\sum_{\l'\in L}m_{(d)}(\l') + \gamma}\ +\ \xi
\label{eq:stable}
\end{eqnarray}
with $\gamma = \epsilon\cdot sgn\left( \sum_{\l'\in L}m_{(d)}(\l') \right)$,  $sgn(y) = 1$ if $y\geq0$ and $-1$ else.
An obvious drawback of this approach is, however, that it violates the conservation constraints set by LRP, as varying amounts of relevance can be absorbed or generated by $\epsilon$.

\subsection{Absolute Value Decomposition}
\label{sec:abs}

Our second alternative to resolve the issue of mapping weight cancellation is to only consider the absolute of a descriptor's mapping contribution, effectively removing all cases of division by zero not already covered by Equation \ref{eq:R2decmop}. As a further benefit, no additional parameters need to be defined and no relevance is lost as in Eq.~\ref{eq:stable}
\begin{eqnarray}
	R_{\l}^{(2)} = & \sum\limits_{d\not\in Z(\x)} R_d^{(3)} \frac{|m_{(d)}(\l)|}{\sum_{\l'\in L}|m_{(d)}(\l')|}\ +\ \xi
\end{eqnarray}
The drawback of this method is the loss of information encoded in the mapping signs.

\section{Measuring the Quality of a Heatmap}
\label{sec:alternatives_comparison}

With \emph{pixel flipping} a procedure has been introduced in \cite{BacBinMonKlaMueSam15}, which demonstrated that the heatmaps computed for the MNIST \cite{lecun1998mnist} data set are able to successfully identify the properties of the input image determining the decision of the classifier. The idea behind this data deterioration approach guided by relevance scores has then been extended in \cite{DBLP:journals/corr/SamekBMBM15} to an evaluation procedure to compare alternatives for LRP decomposition and image perturbation approaches applied to a deep neural network classifier.

Assessing the quality of a heatmap using image deterioration guided by the heatmap itself follows the semantic behind LRP: Let us assume a classifier which predicts a learned target if $f(\x) > 0$. The application of LRP to $f(\cdot)$ with an input point $\x$ should then attribute (the highest) positive values $R^{(k)}_i$ to components of $\x$ which are (most) important to the positive decision of $f(\cdot)$. By intuition,   removing that information should cause the predictor output to become less positive. Considering the values $R^{(k)}_i$ as a \emph{ranking} of  importance of components of the input , e.g. the most important components receive the largest proportions of relevance, then this ranking is optimal if the output of $f(\cdot)$ decreases faster than any other ranking with ongoing removal of input information with a removal order $\vec{o}$ determined by ordering the values $R^{(k)}_i$ descendingly. Since different decomposition methods (e.g. randomly picked component orders) can produce different orderings of importance, an algorithm of relevance-guided data deterioration can be used to compare how well each decomposition variant explains the reasoning of the classifier. We formally define this deterioration algorithm as
\begin{align}
\x^{(0)}_{\MoRF} & = \x \nonumber\\
\forall ~ 1\leq i \leq I ~ : ~ \x^{(i)}_\MoRF & = g\left(\x^{(i-1)}_\MoRF, \vec{o}_i \right)
\label{eq:A}
\end{align}
with $g(\cdot)$ being a perturbation function to remove or exchange data, $\x_\MoRF^{(i)}$ representing a data point after $i$ steps of removal of the {\bf mo}st {\bf r}elevant information {\bf f}irst, determined by an ordering $\vec{o}$, have been performed. We compute a measure of quality of a heatmap $A$ by comparing the prediction $f(\x)$ to the prediction on the altered input $\x_\MoRF^{(i)}$ and then computing the area between those two curves by integrating over the points of measurement:
\begin{align}
A = \frac{1}{I}\sum\limits_{i=1}^I\left( f(\x ) - f( \x ^{(i)}_\MoRF ) \right)
\end{align}
The value $A$ increases with the reaction of the classifier to a change in its input and therefore higher values for $A$ encode a better representation of the classifier decision in terms of back propagated relevance.

\subsection{Local Feature Replacement}
\label{sec:lfreplace}

Previously used variants of pixel flipping have -- hence the name -- operated on pixel level exclusively to evaluate heatmaps for neural network type classifiers. Those classifiers -- namely deep neural networks -- directly received the pixel values of an image as inputs and the application of LRP produce pixel-accurate heatmaps. While \cite{BacBinMonKlaMueSam15} exchanges almost binary pixel values of the 28$\times$28 pixel sized MNIST pixels by inverting the state of each pixel,  \cite{DBLP:journals/corr/SamekBMBM15} studies different image perturbation strategies on 227$\times$227 pixel large color images showing photographic scenes and structures. Here, questions such as the number of pixels to exchange, their spatial grouping and how to best replace those pixels have been raised and the overall complexity of the problem of finding the \emph{right} perturbation strategy is discussed.

When a feature extraction pipeline (in the classical sense: dense local feature extraction $\rightarrow$ mapping $\rightarrow$ pooling) is part of the predictor operating on images
we face a set of distinct problems. For example, exchanging one pixel causes  a recalculation and mapping of all local descriptors covering that pixel in order to measure the effect to the predictor output. Assuming a meaningful heatmap, pixel positions with leading relevance scores are in general not sparsely scattered all over the image area but rather grouped in the same image area over the extend of a patch of neighbouring pixels.
This is even more so true due to the computation of pixel relevance scores $R^{(1)}_p$.
The same local descriptors are very likely to be recalculated over and over again, even though the change in a single pixel will not have much of an effect. This causes a considerable computational cost to assess a heatmap, especially when local descriptors are sampled densely and at multiple scales.
An obvious solution to this problem is to exchange a group of pixels at a time, yet this raises the questions of how many pixels to exchange, due to which grouping and with what replacement strategy? Removing single pixels is out of the question, since this would effectively damage the integrity of the image itself.
Also the type of local descriptor needs to be considered, e.g. do SIFT features encode shape information well.  By blurring an area important structural information might be removed, and a bias towards another class (or even the image's true class) might be introduced, by the blurred area resembling e.g. sky, water or cloth, interfering with the evaluation for classes \emph{aeroplane}, \emph{boat} and \emph{person}, respectively.
Vice versa, setting selected pixels to random color values might create high contrast gradients in the image, affecting the evaluation of certain object classes identifying themselves via sharp and well pronounced edges.

We aim to avoid all those problems by not exchanging pixel values, but local descriptors instead. Firstly, we assume the descriptors forming the orderless descriptor set $L$ serving as an intermediate image representation to be the result independent events which can also be exchanged one at a time without affecting each other's meaning. This is not the case when individual pixels are exchanged.
Secondly, in contrast to pixel scores $R^{(1)}_p$ the local feature scores $R^{(2)}_{\l}$ are also the direct result of the relevance decompositions of the mapping function $m_d(\l)$ of which we wish to compare the alternative options. Se can use the GMM $\lambda$
representing the distribution of local descriptors as a generative model to draw (already dimensionality-reduced) replacements for the features to be exchanged, resulting in believable data points $\x^{(i)}_{\MoRF}$ close to the data manifold. 
What remains to do is to create an ordering of local descriptors to replace and to update the FV $\x$. Algorithm \ref{alg:lfreplace} outlines this perturbation and evaluation strategy. Note, that the transformation $\Phi(\cdot)$ is applied to $\x^{(i)}_{\MoRF}$ before feeding it as an input to $f(\cdot)$ to compute $A$ in Equation \ref{eq:A}.
\begin{algorithm}
\caption{Local Feature Resampling}
\begin{algorithmic}[1]
\State {\bf Input:} local features $L$, GMM $\lambda$, number of replacements $I$, replacement order $\vec{o}$
\State $\x^{(0)}_{\MoRF} \leftarrow \frac{1}{|L|}\sum_{\l\in L} \Psi_{\lambda}(\l)$
\For{$i$ from 1 to $I$}
\State $\l' \leftarrow $ feature to replace $L(\vec{o_i})$ 
\State $\l_{\lambda} \leftarrow \lambda.\text{generate()}$
\State $\x^{(i)}_{\MoRF} \leftarrow \x^{(i-1)}_{\MoRF} + \frac{1}{|L|}\Psi_{\lambda}(\l_{\lambda}) - \frac{1}{|L|}\Psi_{\lambda}(\l')$ 
\EndFor
\end{algorithmic}
\label{alg:lfreplace}
\end{algorithm}
To measure the impact of Algorithm \ref{alg:lfreplace} when applied to an image, we replace the first $10,000$ (wrt to the ordering $\vec{o}$) local descriptors extracted from the image in batches of $100$, resulting in $100$ points of measurement. As a rough estimate, from about $15,000$ up to $100,000$ features are computed for each image of the PASCAL VOC 2007 test set using the dense sampling approach of the encoding evaluation toolbox \cite{chatfield2011devil}, with $69,000$ local features being sampled per image on average.


To reduce random effects influencing the measurement process, we repeat the experiment five times, in total recording 28870 image perturbation traces on all true positive predictions \footnote{5774 TP predictions in total over all 20 classes with 5 repetitions for each case. Single images may show multiple and correctly detected object classes at once.} after optimizing the prediction threshold wrt to the EER measure.


\section{Details on the Neural Network Retraining}

The starting point was the BVLC reference caffe net as provided with the caffe package \cite{jia2014caffe}. Training mode was multi-label training instead of the usual competitive multi-class training because for PASCAL VOC multiple classes can be present in one image. The training criterion was the sum of hingelosses over all 20 classes in the Pascal VOC data set. Note that this required to use a customized image data layer and a customized hinge loss layer.

One general problem for training and testing is the question how to score and image and what data to use for training. A second problem is how to generate patches matching the quadratic receptive field size from non-quadratic images. One general approach to generate non-quadratric images is to ignore the aspect ratio and to use warping in order to transform a non-quadratic patch into a quadratic one such as in \cite{DBLP:journals/corr/GirshickDDM13}. 

In order to have maximal comparability to Fisher vectors which do not use warping and process an image as a whole we decided for a simpler setup which is close to the setup used by Fisher vectors and which preserved the aspect ratio of patches used during training and testing, irrespective of the fact that other setups may have resulted in somewhat higher performance of the neural network.

\subsubsection*{Training data:}

As we were interested in training a setup such that the neural network is able to use context, we refrained from training the network with image patches around the scale of a bounding box and smaller. We decided not to use the information about object bounding boxes for generating training data because for Fisher vectors this information was not used for generating training data. 
Instead each image was rescaled such that the largest side had 256 pixels. The smaller side was padded at its boundaries by the nearest pixel. From this modified image 4 edge and one center crop was taken. After mirroring the image, this was repeated. This resulted in 10 images per training image. This is a compromise to ensure a sufficiently large sample size for retraining, as it is known that neural networks excel typically at higher training sample sizes.

\subsubsection*{Testing data:}

The heatmap was computed using one center crop only.

As for measurement of mean average precision, results depend on how to score one image at test time. Note that it is common to compute an average score over many crops of the image.

Resizing the largest side of the image to 256 pixels and using the $227 \times 227$center crop only for each image resulted in the 72.12 mAP reported in the main paper. Note that this setup corresponded to the setup used for computing the heatmaps, so this was used for the sake of comparability. 

Using a different test strategy, namely resizing the smallest side of the image to 256 pixels, then computing an average over a sliding window with stride of 20 pixels, resulted in an increase of ranking performance to 75.9 mAP. This strategy used merely $12$ - $35$ test patches per image. Using approaches with several hundred test windows, such as 500 windows in \cite{DBLP:conf/cvpr/OquabBLS14} would probably have resulted in much higher mAP scores, however we did not consider a higher score relevant for the main message of the paper focused on context usage. In view of the considerably increased computation time for such approaches we refrained from them.

\end{document}